\documentclass[letterpaper, 10 pt, conference]{ieeeconf}  % Comment this line out if you need a4paper

\IEEEoverridecommandlockouts                              % This command is only needed if 
\usepackage{adjustbox}

\usepackage{multicol}
\usepackage{caption}
\usepackage{blindtext}
\usepackage{graphicx}
\usepackage{hyperref}
\usepackage{amsmath}
\usepackage{amssymb}
\usepackage{tabularx}
\usepackage{subcaption}
\usepackage{xcolor}
\usepackage[acronym]{glossaries}
\usepackage{wrapfig,lipsum,booktabs}
\usepackage{siunitx}
\usepackage{float}
\DeclareMathOperator{\arctantwo}{arctan2}

\newcommand{\predacc}{89.9\%}

\newcommand{\precision}{0.842}
\newcommand{\recall}{0.888}
\newcommand{\fscore}{0.865}
\newcommand{\aucscore}{0.96}

\usepackage{multirow}
\usepackage[normalem]{ulem}
\useunder{\uline}{\ul}{}

\newacronym{dl}{DL}{Deep Learning}
\newacronym{gat}{GAT}{Graph Attention Network}
\newacronym{gats}{GATs}{Graph Attention Networks}
\newacronym{gnns}{GNNs}{Graph Neural Networks}
\newacronym{gnn}{GNN}{Graph Neural Network}
\newacronym{ccg}{CCG}{Corner Case Generation}
\newacronym{cnns}{CNNs}{Convolutional Neural Networks}
\newacronym{hgnn}{HGNN}{Heterogeneous Graph Neural Network}

\newacronym{gcns}{GCNs}{Graph Convolutional Networks}
\newacronym{gcn}{GCN}{Graph Convolutional Network}
\newacronym{rgcn}{RGCN}{Relational Data with Graph Evolutionary Networks}
\newacronym{hin}{HIN}{Heterogeneous Information Network}
\newacronym{hins}{HINs}{Heterogeneous Information Networks}
\newacronym{auc}{AUC}{Area Under Curve}
\newacronym{roc}{ROC}{Receiver Operating Characteristic}
\newacronym{tpr}{TPR}{True Positive Rate}
\newacronym{fpr}{FPR}{False Positive Rate}
\newacronym{elu}{ELU}{Exponential Linear Unit}
\newacronym{npc}{NPC}{Non-Player Character}
\newacronym{sgg}{SGG}{Scene Graph Generation}
\newacronym{bp}{BP}{back-propagation}
\newacronym{hgn}{HGN}{Heterogeneous Graph Attention Network}
\newacronym{pyg}{PyG}{PyTorch Geometric}
\newacronym{mlp}{MLP}{Multi-Layer Perceptron}
\newacronym{mlps}{MLPs}{Multi-Layer Perceptrons}
\newacronym{adam}{Adam}{Adaptive Moment Estimation}
\newacronym{nhtsa}{NHTSA}{National Highway Traffic Safety Administration}
\newacronym{mlms}{MLMs}{Masked Language Models}
\newacronym{gans}{GANs}{Generative Adversarial Nets}
\newacronym{vae}{VAE}{Variational Autoencoders}
\newacronym{rl}{RL}{Reinforcement Learning}

\usepackage{xcolor}

\usepackage[hang,flushmargin]{footmisc}

\begin{document}

\title{\LARGE \bf
% Watch where you are not going:
CC-SGG: Corner Case Scenario Generation using Learned Scene Graphs
}
\author{George Drayson$^*$, Efimia Panagiotaki$^*$, Daniel Omeiza, Lars Kunze \\
Cognitive Robotics Group, Oxford Robotics Institute, \\
Department of Engineering Science, University of Oxford, UK\\
\texttt{george.drayson.23@ucl.ac.uk, \{efimia,danielomeiza,lars\}@robots.ox.ac.uk}\thanks{$^*$ These authors contributed equally to this work. \vspace{0.15cm}}
\thanks{This work was supported by a DeepMind Engineering Science Scholarship, the EPSRC project RAILS (grant reference: EP/W011344/1), and the Oxford Robotics Institute research project RobotCycle.
}
}
\maketitle

%%%%%%%%%%%%%%%%%%%%%%%%%%%%%%%%%%%%%%%%%%%%%%%%%%%%%%%%%%%%%%%%%%%%%%%%%%%%%%%%
\begin{abstract}
Corner case scenarios are an essential tool for testing and validating the safety of autonomous vehicles (AVs).
As these scenarios are often insufficiently present in naturalistic driving datasets, augmenting the data with synthetic corner cases greatly enhances the safe operation of AVs in unique situations. 
However, the generation of synthetic, yet realistic, corner cases poses a significant challenge. 
In this work, we introduce a novel approach based on \glspl{hgnn} to transform regular driving scenarios into corner cases. To achieve this, we first generate concise representations of regular driving scenes as scene graphs, minimally manipulating their structure and properties. Our model then learns to perturb those graphs to generate corner cases using attention and triple embeddings. The input and perturbed graphs are then imported back into simulation to generate corner case scenarios.
% used to create corner case scenarios in simulation. %We exploit the expressive power of graphs to learn embedding representations of road users and infrastructure, encoding their respective relationships while learning to predict links corresponding to corner case scene graphs. 
Our model successfully learned to produce corner cases from input scene graphs, achieving $\mathbf{\predacc}$ prediction accuracy on our testing dataset.
%, demonstrating the effectiveness of our approach. 
We further validate the generated scenarios on baseline autonomous driving methods, demonstrating our model's ability to effectively create critical situations for the baselines.

% showcasing that our model successfully generates critical situations, resulting in a notable decrease in the AD systems' accuracy.

% significantly reducing their accuracy.

% that our model successfully generates critical situations for AD systems, significantly reducing their planning accuracy.

% the suitability and practicality of our generated scenarios on various state-of-the art systems, proving the necessity of such scenarios for validating the safety of AD systems.

% the suitability of our generated scenarios in simulation, generating critical scenarios on state-of-the-art systems.

% We further demonstrate that importing the generated scenarios into a simulator 
%Our method can be split into three main parts; generation of heterogeneous scene graphs from simulated scenarios, encoding of graph features using attention, and triples learning for link prediction. Our model successfully learned to generate corner cases from input scene graphs, achieving $\mathbf{\predacc}$ prediction accuracy on our testing dataset, demonstrating the effectiveness of our approach. \ep{TODO: Review abstract}
\end{abstract}

%%%%%%%%%%%%%%%%%%%%%%%%%%%%%%%%%%%%%%%%%%%%%%%%%%%%%%%%%%%%%%%%%%%%%%%%%%%%%%%%
\section{Introduction}
    To enable successful and wide integration of autonomous driving (AD), vehicles need to operate safely in complex and unexpected traffic situations. Progress to-date in the development of self-driving technology can be attributed in part to the increased availability of diverse datasets for training and testing of AD systems \cite{liu2021survey}. However, a well known weakness of most autonomous systems, and particularly those based on deep learning models, is their lack of robustness in handling corner case situations \cite{DLsurvey}. 
    A corner, or \emph{edge}, case refers to an infrequent or unique scenario that lies at the extreme limits of a system's expected behaviour creating a critical situation. Extensive testing in rare yet potentially dangerous scenarios ensures the vehicles' safe operation in unexpected and challenging circumstances \cite{sun2021corner}.

In real-world datasets, the complexity of traffic scenarios adheres to a \emph{long-tail} distribution due to the scarcity of corner cases, thereby creating substantial implications for the safe and reliable operation of self-driving systems 
\cite{zhou2022longtail, makansi2021exposing, zhang2023deep}. Long-tail data usually contain either high-level traffic scenes and scenarios depicting infrequent and challenging traffic patterns, or low-level adverse training data for model learning. The former are commonly considered the most complex to detect as they involve sequential patterns and contextual anomalies \cite{breitenstein2020systematization, breitenstein2021corner}. Corner-case-specific datasets in existing literature mainly focus on object-, domain-, and pixel- level corner cases \cite{breitenstein2020systematization}. Existing datasets containing scenario- and scene- level corner cases \cite{pinggera2016lost, Zendel2018WildDashC} do not include the required temporal information for scenario-level corner case methods and thus existing approaches rely on custom-made real-world or simulation-based artificial datasets.
% Lost and Found \cite{pinggera2016lost} and WildDash \cite{Zendel2018WildDashC} are the only datasets containing also scenario- and scene- level corner cases. However, as they do not include temporal information they cannot easily be used by scenario-level corner case methods. 
% %and thus current literature relies on custom-made real-world or simulation-based artificial datasets.

In this work, we address the problem of long-tail distributions
%of corner case traffic scenes and scenarios 
in AD datasets by proposing a novel methodology that learns to generate corner case scenarios directly from regular driving scenes. We first generate concise scene graph representations of key semantic elements and relationships in our custom simulation-based dataset. Our model then encodes the graph entities, leveraging their heterogeneity, and perturbs the input regular scene graphs to generate corresponding corner case graphs. Given these regular and corner case scenes, we then generate corner case scenarios in the CARLA AD Simulator.

%learns embedding representations of the heterogeneous scene graph entities and perturbs the input to generate a corresponding corner case graph. Given the 

% In our approach, the model learns embedding representations of heterogeneous scene graph entities and then perturbs the input to generate a corresponding corner case graph. To generate the graphs we extract key semantic entities and relationships from the scenes in our custom dataset. \ep{rephrase/fix this} The input scene graphs are extracted from a list of simulated scenarios in a custom dataset, containing regular and corner case scenes. To build the graphs, we establish an \emph{ontology} capturing domain knowledge from real traffic scenario topology, as well as a \emph{grammar} to define the rules governing the graph structure and morphology. This ensures the plausibility of the generated corner cases while rooting the graphs to real traffic scenarios. 
Our key contributions are summarised as follows:
\begin{itemize}
    \item We propose a novel end-to-end scene graph based methodology for generating corner case scenarios in simulation by perturbing regular driving scenes;
    \item We extend the attention and triple embeddings learning modules of \glspl{hgnn} to encode key entity attributes and predict links corresponding to corner case graphs;
    \item We validate our generated scenarios across five benchmarks demonstrating the effectiveness of our approach in creating critical situations for AD systems.
\end{itemize}
% The novelty of our work lies in our unique approach to address the challenge of generating corner case scenarios using HGNNs perturbing regular driving scenes. We have designed a novel model and pipeline for link prediction and feature encoding. In our model, we exploit attention to encode edge attributes directly into node feature vectors. Furthermore, we introduce a novel triple learning embedding representation methodology, extending link prediction methods. We approach this problem as a supervised classification task for link prediction, using labelled extended scene graphs as ground truth, which is a novel approach for corner case generation. With this work, we leverage the power of graph-based learning for corner case generation, aiming to establish a foundational benchmark for future research endeavours.
%\ep{review}.To benchmark our results, we extensively validate the generated scenarios across five different planning algorithms, demonstrating their effectiveness in generating challenging situations. 
Quantitative and qualitative results affirm that our model successfully learns to generate corner cases from regular scene graphs, producing challenging scenarios in simulation for autonomous driving. 
\begin{figure*}[ht]
\centering
  \includegraphics[width=\textwidth]{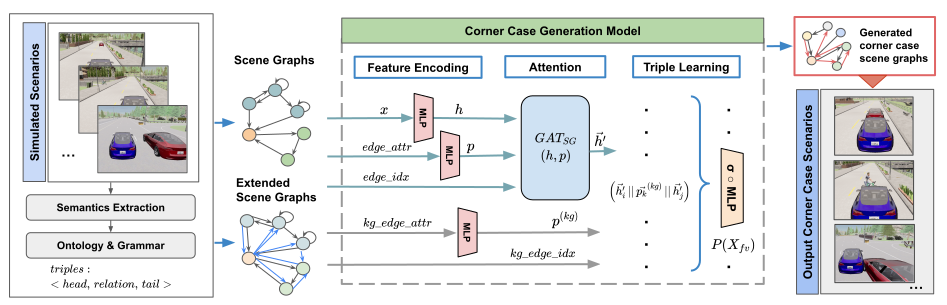}
   \caption{Overview of our proposed methodology. We first extract semantics and relationships from simulated scenarios to generate heterogeneous scene graphs following a pre-defined generic knowledge graph ontology and grammar. Our model then encodes graph attributes generating concise feature representations to predict links corresponding to corner case scene graphs. The output of the model is a perturbed scene graph of an input \emph{regular driving} graph. Scene graphs of the predicted corner case and the input regular driving graph are then used to generate corner case scenarios in simulation. }
   %We first generate heterogeneous scene graphs extracting semantics and relationships from simulated scenarios. Following a pre-defined generic knowledge graph ontology and grammar, we extract extended scene graphs to assist our link prediction module. We then encode graph attributes
  
  % . To predict links corresponding to corner case graphs, we augment the input following a pre-defined generic knowledge graph ontology and grammar. Our  
  %We generate a dataset containing heterogeneous scene graphs extracted from simulated scenarios. To predict links corresponding to corner case graphs, we augment the input following a pre-defined generic knowledge graph ontology and grammar. To encode the graph features, we propose a novel attention module and triplet learning process. The output of the model is a perturbed scene graph of the input, corresponding to a corner case. Our model takes as input a scene graph and its extended version. It then generates feature vector representations for each graph entity by encoding its neighborhood using sets of \gls{mlps} followed by a graph attention module $GAT_{SG}$ to generate triple representations. After encoding the feature vectors of the triples, the model learns to predict an array of probabilities corresponding to the existence of links in the extended scene graph.
  \label{fig:overview}
  \vspace{-0.3cm}
\end{figure*}

\section{Related Work}\label{sec:related_work}
Multiple studies address the task of corner case generation, aiming to construct either task-specific adversarial data or high-level traffic scenarios for testing and evaluation of autonomous systems.

\textbf{Task-specific data generation.} 
Recent approaches leverage deep learning methods to generate adversarial data.
STRIVE \cite{rempe2022strive} uses data-driven motion priors on a generative model \cite{goodfellow2014gans} to create plausible critical scenarios for the AD planner. Causal BERT \cite{resnick2022bert} encodes scene graph representations and then uses \gls{mlms} to semantically perturb the input for object detection. For collision avoidance, a common practice is learning to minimise the driveable area to generate challenging scenarios \cite{althoff, klischat}. Methods using risk directly as training signal for generative or \gls{rl} models have been introduced for tasks like decision making \cite{multimodal, Ding2020LearningTC} and lane changing \cite{adversarialeval}. As these approaches aim to generate corner cases for a particular task, they cannot be generalised easily.

% their suitability for end-to-end AD systems is often questioned.
% As these approaches are not based on or linked to plausible scenarios, their practical value is often questioned.

\textbf{Traffic scenario generation.} In the domain of scenario corner case generation, various learned and rule-based approaches have been introduced. 
AdvSim \cite{Wang2021advsim} directly changes the input lidar pointcloud data by removing and adding simulation-generated actors. TrafficGen \cite{feng2022trafficgen} learns to augment scenarios in HD maps using an encoder-decoder architecture and then transfers them back in simulation. These methods rely on artificially generated agents and thus their produced scenarios lack plausibility. 
% Aiming to produce realistic synthetic datasets, Meta-Sim \cite{kar2019metasim} and Meta-Sim2 \cite{devaranjan2020metasim2} were introduced. These models perturb scene graphs to recreate driving scenarios from real-world datasets, in simulation. Even though these methods produce satisfactory results, they still rely on manual searches to detect corner cases. 
Rule-based methods generate scenarios directly using domain knowledge of traffic scenes represented as an \emph{ontology}, without requiring specific scene representations \cite{kluck2018using, bogdoll2022ontology}. However, as these scenarios are generated solely from an abstract ontology, they do not link to real traffic data, and thus, lack plausibility.

\textbf{Heterogeneous graph learning.} Heterogeneous graphs contain different types of entities and their respective relationships, modelling complex and semantically-rich data for downstream learning tasks \cite{hgnn}. Recent approaches address the challenge of modelling multiple types of nodes and relationships using meta paths to generate vector representations \cite{shi2016survey, li2018link, wang2019heterogeneous}. However, it is frequently observed that these methods tend to lose significant topological and featural information inherent in the graphs \cite{Hong}. Methods extending classical \gls{gnn} models have also been introduced, exploiting attention to embed structural and semantic information \cite{Hong}, as well as convolutions to introduce relation-specific transformations into the feature vectors \cite{schlichtkrull2018modeling}.

Our approach focuses on the task of corner case traffic scenario generation using heterogeneous graph embedding representations. 
% It relies on scene graphs built from actual traffic data, constrained by an ontology and grammar, to ensure plausibility. 
We exploit the expressive power of scene graphs to encode traffic agents along with their state and neighbouring relationships. Our model extends current \gls{hgnn} methods by encoding edge attributes directly inside node feature vectors using attention and relying on triple embedding representations for predicting links corresponding to corner cases.
% capturing the complexity of the traffic scenes encoding graph attributes using attention and triple learning to predict links corresponding to corner case scene graphs. 
Regular and perturbed scenes are then imported into the simulation to generate safety-critical scenarios for AD systems.

% . Extending current \gls{hgnn} methods, we design a model that encodes edge attributes directly inside node feature vectors, using attention. The model then generates triple embedding representations and learns to predict links corresponding to corner cases. Our method generates informative and concise feature representations that capture the complexity of the traffic scenes. \ep{make this smaller perhaps and talk about how we add the scenarios back in simulation}
% Our low-cost scenario generation model achieves satisfactory results without requiring extensive training datasets.
%===============================================================================

%\section{Corner Case Scenario Generation}
\section{Overview}
\label{sec:methodology}
In our approach, visualised in Figure \ref{fig:overview}, we first generate a dataset in simulation containing a set of pre-corner-case scenarios, either pre-crash or near-misses. The semantic elements from distinct frames in each scenario are then extracted to generate corresponding scene graphs, discussed in \ref{sec:sg_construction}. We categorise each scene graph accordingly to generated sets of \emph{regular driving graphs} and \emph{corner case graphs}. The latter serves as both the ground truth and supervisory signal. The model learns to perturb input regular driving graphs, generating associated corner cases, described in \ref{sec:edge_case_learning}. Using the regular and the generated scene graphs we then create corner case scenarios in CARLA. 

\textbf{Problem Definition and Notations}
\label{subsec:problemdef}
Let $F_{t}$ be a frame representing a traffic scene at discrete timestamp $t$. Each scenario consists of $N$ consecutive frames, associated with a corresponding timestamp. The last frame $F_N$ represents a corner case scene. For each $F_{t}$ we generate a scene graph $\mathcal{G}_{t} = (\mathcal{V}_{t}, \mathcal{E}_{t})$; the nodes $\mathcal{V}_{t}$ correspond to key dynamic and static actors in the scene and the edges $\mathcal{E}_{t}$ correspond to relationships between them. Each node $n_i \in \mathcal{V}$ depicts a discrete semantic category (e.g., ego vehicle, road, pedestrian, cyclist) associated with a node feature representation ${h}_i$. Similarly, each edge $e_i \in \mathcal{E}$ represents a semantic relationship (e.g., distance, topology) with an associated feature representation ${p}_i$. Naturally, $\mathcal{G}_N$ corresponds to the graph derived from the corner case frame. 
In our proposed method, the aim is to learn to perturb a query graph $\mathcal{G}_{t}$, $t \in \{0, 1, \dots, N-1\}$ to generate a corner case graph $\mathcal{G}_N$. 
Given the initial and target locations of the non-ego-vehicle agents -- also referred to as \emph{adversaries} -- in the respective query and generated graphs, we can then create a corner case scenario in simulation. 

\section{Scene Graph Construction}
\label{sec:sg_construction}

\subsection{Dataset}
To generate our dataset, we developed a set of scenarios in the ASAM OpenScenario standard \cite{ASAM} for implementation in CARLA. In each run, the \emph{ego vehicle} is controlled by the user and the behaviour of all adversaries is defined in the respective OpenScenario file. Our dataset consists of the scenarios identified in the CARLA AD Challenge \cite{challenge}, similar to \cite{resnick2022bert, klischat, adversarialeval, bogdoll2022ontology}, as well as scenarios from the NHTSA pre-crash topology at the General Estimates System (GES) crash database for Crash Avoidance Research \cite{NHTSA}.
To increase the diversity of our dataset, we also developed scenarios based around motorway driving such as merging and lane changing. 
% In each run, the \emph{ego vehicle} is controlled by the user and the behaviour of all adversaries is defined in the respective OpenScenario file. \ep{}

% To generate our dataset, we create a series of scenarios in CARLA AD simulator \cite{carla}. The scenarios were developed following the ASAM OpenScenario standard \cite{ASAM}, commonly used to describe complex traffic scenarios in driving simulators, and were then loaded in CARLA using ScenarioRunner\footnote{ScenarioRunner: \url{https://carla-scenariorunner.readthedocs.io}}. In each run, the \emph{ego vehicle} is controlled by the user and the behaviour of all adversaries is defined in the respective OpenScenario file. \ep{this can be removed}

% To make our scenarios as close to real-world driving as possible, we follow a pre-crash scenario topology defined by the \gls{nhtsa}, at the General Estimates System (GES) crash database for Crash Avoidance Research \cite{NHTSA}. To generate a larger dataset of corner cases, we developed a series of custom scenarios based around motorway driving corner cases, such as merging and lane changing. 

\subsection{Scene Graph Extraction} 
\textbf{Ontology and grammar.} To generate concise graph representations, we establish an \emph{ontology} and a \emph{grammar} capturing domain knowledge from real traffic scenario topology, as depicted in Table \ref{table:triples}.
%, as well as a \emph{grammar} to define the rules governing the graph structure and morphology. This ensures the plausibility of the generated corner cases while rooting the graphs to real traffic scenarios. 
In the context of scene graphs, the ontology defines a class hierarchy of objects and relationships that can exist in a graph. The grammar defines a set of rules governing the graph structure and morphology, ensuring consistency in the representations from the dataset. The grammar guides the generation of ontology triples in the form of $\langle \text{head}, \text{relation}, \text{tail} \rangle$, which are then used to generate knowledge graph representations. 
%The ontology and grammar we defined for our scene graphs are depicted in Table \ref{table:triples}.

\begin{table*}[]
\vspace{0.2cm}
\centering
\resizebox{0.9\textwidth}{!}{
\renewcommand{\arraystretch}{1.1}
\begin{tabular}{llllll}
\multicolumn{1}{l|}{\textbf{Head Actor}} & \multicolumn{1}{l|}{\textbf{Ego vehicle (E)}} & \multicolumn{1}{l|}{\textbf{Lane (L)}} & \multicolumn{1}{l|}{\textbf{Pavement (Pa)}} & \multicolumn{1}{l|}{\textbf{Shoulder (S)}} & \multicolumn{1}{l}{\textbf{Road (R)}}  \\ \hline
\multicolumn{1}{l|}{\textbf{\begin{tabular}[c]{@{}l@{}}Self-\\ Relation\end{tabular}}} & \multicolumn{1}{l|}{\begin{tabular}[c]{@{}l@{}}location $(x,y)$\\ velocity $(v_x, v_y)$\\ braking (binary)\end{tabular}}& \multicolumn{1}{l|}{--}& \multicolumn{1}{l|}{--}& \multicolumn{1}{l|}{--}& \multicolumn{1}{l}{--}\\ \hline
\multicolumn{1}{l|}{\begin{tabular}[c]{@{}l@{}}\textbf{Relation}\\ $\rightarrow$\\ \textbf{Tail Actor}\end{tabular}}      & \multicolumn{1}{l|}{\begin{tabular}[c]{@{}l@{}}isIn \\  $\rightarrow$ \{L, Pa, S\}\\distance  
\\  $\rightarrow$ \{C, B, P, O, T, S\}\end{tabular}}& \multicolumn{1}{l|}{\begin{tabular}[c]{@{}l@{}}isIn \\  $\rightarrow$ \{R\}\end{tabular}} & \multicolumn{1}{l|}{\begin{tabular}[c]{@{}l@{}}isIn \\  $\rightarrow$ \{R\}\end{tabular}} & \multicolumn{1}{l|}{\begin{tabular}[c]{@{}l@{}}isIn \\  $\rightarrow$ \{R\}\end{tabular}} & \multicolumn{1}{l}{--} \\ 
\multicolumn{6}{l}{} \\ 
\multicolumn{1}{l|}{\textbf{Head Actor}}          & \multicolumn{1}{l|}{\textbf{Car (C)}}& \multicolumn{1}{l|}{\textbf{Bicycle (B)}} & \multicolumn{1}{l|}{\textbf{Pedestrian (P)}} & \multicolumn{1}{l|}{\textbf{Traffic Light (T)}} & \multicolumn{1}{l}{\textbf{Object (O)}}  \\ \hline
\multicolumn{1}{l|}{\textbf{\begin{tabular}[c]{@{}l@{}}Self-\\ Relation\end{tabular}}} & \multicolumn{1}{l|}{\begin{tabular}[c]{@{}l@{}}location $(x,y)$\\ velocity $(v_x, v_y)$\\ braking (binary)\end{tabular}}     & \multicolumn{1}{l|}{\begin{tabular}[c]{@{}l@{}}location $(x,y)$\\ velocity $(v_x, v_y)$\end{tabular}}   & \multicolumn{1}{l|}{\begin{tabular}[c]{@{}l@{}}location $(x,y)$\\ velocity $(v_x, v_y)$\end{tabular}}  & \multicolumn{1}{l|}{lightState}   & \multicolumn{1}{l}{--}  \\ \hline
\multicolumn{1}{l|}{\begin{tabular}[c]{@{}l@{}}\textbf{Relation}\\ $\rightarrow$\\ \textbf{Tail Actor} \end{tabular}}      & \multicolumn{1}{l|}{\begin{tabular}[c]{@{}l@{}}isIn \\$\rightarrow$ \{L, Pa, S\}\\ distance \\ $\rightarrow$ \{E\}\\ relativePosition  \\ $\rightarrow$ \{E\}\end{tabular}} & \multicolumn{1}{l|}{\begin{tabular}[c]{@{}l@{}}isIn \\ $\rightarrow$ \{L, Pa, S\}\\ distance \\ $\rightarrow$ \{E\}\\ relativePosition  \\ $\rightarrow$ \{E\}\end{tabular}} & \multicolumn{1}{l|}{\begin{tabular}[c]{@{}l@{}}isIn \\$\rightarrow$ \{L, Pa, S\}\\ distance  \\$\rightarrow$ \{E\}\\ relativePosition  \\ $\rightarrow$ \{E\}\end{tabular}} & \multicolumn{1}{l|}{\begin{tabular}[c]{@{}l@{}}isIn \\  $\rightarrow$ \{L, Pa, S\}\end{tabular}} & \multicolumn{1}{l}{\begin{tabular}[c]{@{}l@{}}isIn \\  $\rightarrow$ \{L, Pa, S\}\end{tabular}} \\ 
\multicolumn{6}{l}{} \\ 
\end{tabular}}
\vspace{-0.3cm}
\caption{Ontology of nodes and relations guided by the grammar in the form of triples $\langle \text{Head Actor}, \text{Relation}, \text{Tail Actor} \rangle$. The direction of the edges in the graph is encoded in the triples. Node and edge features are categorical variables while edge attributes are numerical or binary. Self-relations describe self-edge attributes in the form of $\langle \text{Head Actor}, \text{Relation}, \text{Head Actor} \rangle$.}
\label{table:triples}
\vspace{-0.3cm}
\end{table*}

\textbf{Graph Representations.} The scene graphs derive from the actors in each scenario and their respective relationships, following the ontology and grammar in Table \ref{table:triples}. The attributes of the self-edges describe the independent state of each agent in the scene. 
To facilitate the downstream learning task of corner case generation, we discretise edge attributes belonging to non-self-edges, referred to as \emph{cross-edges}, simplifying the task by making the number of possible edges for link prediction finite. 
The proposed method aims to perturb all cross-edges in a graph, taking into account the possibility of edges of multiple relation types existing between nodes. 

\begin{figure}
\vspace{0.3cm}
\centering
\includegraphics[width=0.9\linewidth]{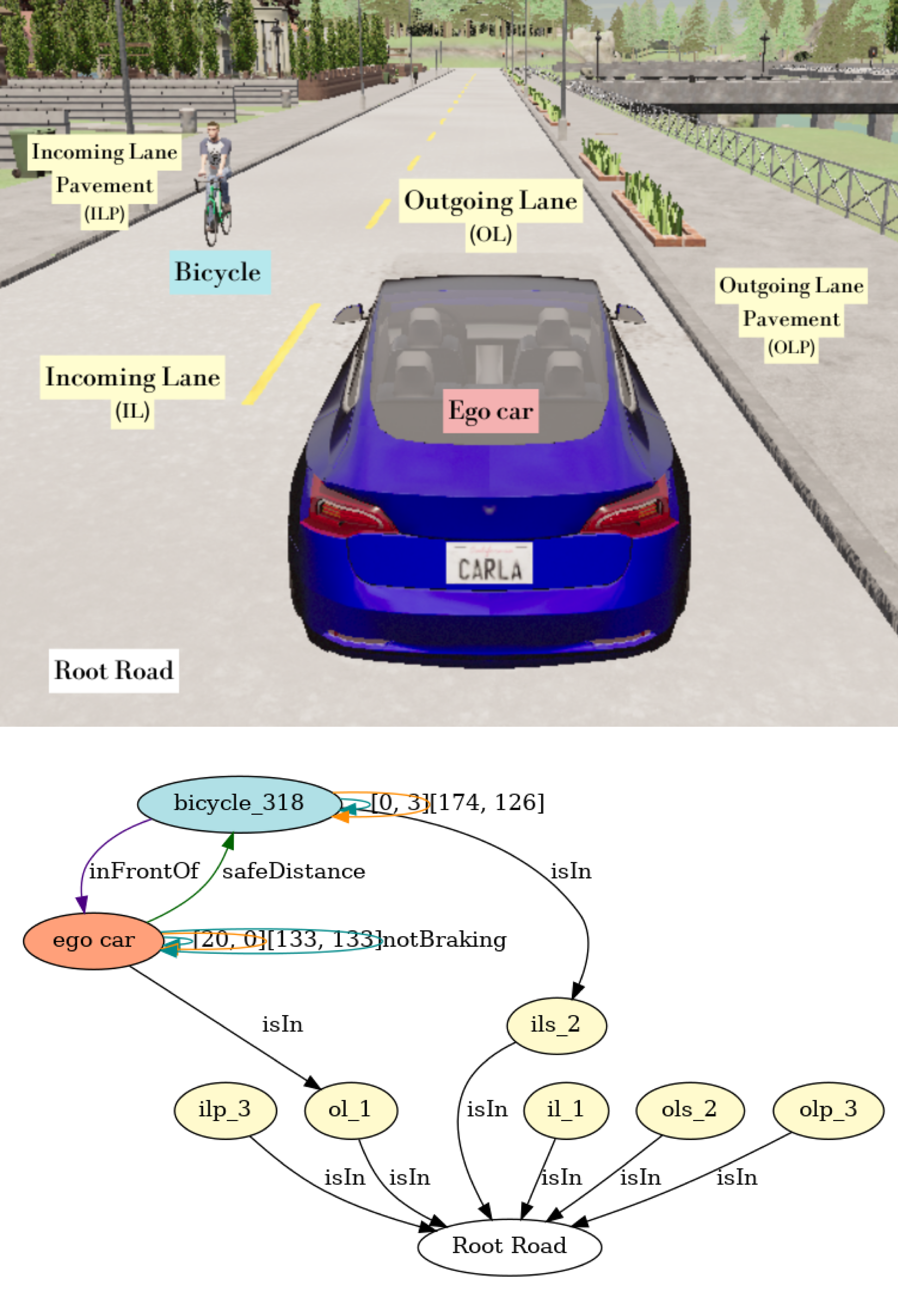}
 \caption{Simulated scenario from \emph{regular driving} (up) and its associated scene graph (down) of a bicycle in front of the ego vehicle travelling in the opposite lane. Self-edge attributes correspond to location $(x=133, y=133)$, speed $(v_x=20 m/s, v_y=0 m/s)$, and braking status. Cross-edge attributes correspond to the spatial relations between actors.}
% The self-edge attributes of the ego vehicle are: location $(x=147, y=133)$, speed $(v_x=10 m/s, v_y=0 m/s)$, and braking status. The `isIn' relation corresponds to the position of each actor w.r.t the lane topology. The `safeDistance' and `inFrontOf' attributes describe the spatial relationship between ego vehicle and bicycle. \ep{review image and text}}
\label{fig:fig2}
\vspace{-0.5cm}
\end{figure}

The relation categories affected by this discretisation are \emph{relativePosition} and \emph{distance}. To discretise the relative position relation w.r.t the coordinate frame of the head actor, we first calculate the distance between the actors positioned in $(x_1, y_1)$ and $(x_2,y_2)$ in the horizontal plane. Then, taking into account the orientation of the tail actor, $\theta_{actor}$, we calculate the relative angle between them, $\delta_{rot}$, based on an odometry model for estimating the motion between two points \cite{stahl}: 
\begin{equation}\label{eq:rotation}
    \delta_{rot} =  \arctantwo({x_2} - {x_1},{y_2} - {y_1}) - \theta_{actor}
\end{equation}
Using $\delta_{rot}$ we determine the quadrant in which the tail actor is placed w.r.t the head actor's coordinate frame, discretising the relative position relation into one of the categories: `inFrontOf', `atRearOf', `toLeftOf', or `toRightOf'.

The distance relation is discretised into `safe' and `unsafe' according to the typical vehicle stopping distances provided by the Highway Code \cite{rule126}. This guide is used to define the `safe' and `unsafe' distance margins between a vehicle and the adversaries.
%===============================================================================
% \textbf{Extended Scene Graphs Generation.} 
\section{Corner Case Generation}
\label{sec:edge_case_learning}

\subsection{Extended Scene Graphs} Our model learns to predict the existence of links between regular scenario entities that correspond to corner case graphs. Recent studies on link prediction rely on brute-force techniques to exhaustively search for single links between node-pairs in graphs \cite{ wang2019heterogeneous, Zong2022}. Our method extends conventional approaches by predicting the existence of multiple links and attributes between nodes.
In our approach, we augment the input according to the entities in the knowledge graphs generated from Table \ref{table:triples}. These graphs contain all possible edges and edge attributes between the nodes present. The model then learns to predict the probability for each edge being an actual edge in the ground truth corner case graph.

\subsection{Link Prediction Model}
\textbf{Graph Feature Embeddings.} Input graph entities are represented by either continuous numerical values or one-hot encodings. To handle the heterogeneity of the graphs, the model needs to encode the input and generate concise feature vector representations. To do so, we employ a series of \gls{mlps},
% to learn embedding feature representations of input features, 
encoding node features $\mathbf{h} = MLP(x) = \{ \vec{h}_1, \vec{h}_2, ..., \vec{h}_N  \}, \vec{h}_i \in \mathbb{R}$, scene graph edge attributes $\mathbf{p} = MLP(edge\_attr) = \{ \vec{p}_1, \vec{p}_2, ..., \vec{p}_M  \}, \vec{p}_i \in \mathbb{R}$, and extended scene graph edge attributes $\mathbf{p^{(kg)}} = MLP(kg\_edge\_attr) = \{ \vec{p}^{(kg)}_1, \vec{p}^{(kg)}_2, ..., \vec{p}^{(kg)}_Q  \}, \vec{p}^{(kg)}_i \in \mathbb{R}$, where $N$, $M$, and $Q$ are the number of nodes, scene graph edges, and extended scene graph edges respectively. Using attention, we aggregate the feature vectors from each query node with its neighboring feature vectors to generate $\mathbf{h^{\prime}} = GAT_{SG}(\mathbf{h}, \mathbf{p}) = \{ \vec{h}^{\prime}_1, \vec{h}^{\prime}_2, ..., \vec{h}^{\prime}_N \}, \vec{h}^{\prime}_i \in \mathbb{R}$. $GAT_{SG}$ is our graph attention module.
\newline

\textbf{Graph Attention Module. }We extend the \gls{gat} \cite{velivckovic2017graph} so that multi-dimensional edge features are also taken into account in the attention module. The attention coefficients, indicating the importance of each node $n_j$ to node $n_i$, are calculated as follows: 
\begin{equation}
     c_{ij} = a(\mathbf{\Theta}\vec{h}_i, \mathbf{\Theta}\vec{h}_j, \mathbf{\Theta}_{p} \vec{p}_{k})
\end{equation}
Where $a$ is the vanilla attention mechanism and $\Theta\in\mathbb{R}^{F'\times F}$ is a weight matrix, with $F$ and $F'$ corresponding to the number of initial and new features in each node respectively. Similar to vanilla-GAT, the coefficients are then passed into a softmax and a LeakyRELU:
% \begin{equation}\label{eq:attention}
    \begin{align}\label{eq:attention}
    \alpha_{i,j} &= softmax(c_{ij}) = \frac{c_{ij}}{\sum_{r \in \mathcal{N}}exp(c_{ir})} \\
   \alpha_{i,j} &=\frac{
e^{\left(\mathrm{LeakyReLU}\left(\mathbf{a}^{\top}
[\mathbf{\Theta}\vec{h}_i \, \Vert \, \mathbf{\Theta}\vec{h}_j
\, \Vert \, \mathbf{\Theta}_{p} \vec{p_k}]\right)\right)}}
{\sum_{r \in \mathcal{N}_i}
e^{\left(\mathrm{LeakyReLU}\left(\mathbf{a}^{\top}
[\mathbf{\Theta}\vec{h}_i \, \Vert \, \mathbf{\Theta}\vec{h}_r
\, \Vert \, \mathbf{\Theta}_{p} \vec{p_m}]\right)\right)}}
    \end{align}
% \end{equation}
These normalised attention coefficients are then used to compute the final output feature vectors for each node from a linear combination of the neighbour and self feature vectors:
\begin{equation}
\vec{h}^{\prime}_i = \alpha_{i,i}\mathbf{\Theta}\vec{h}_{i} +
\sum_{j \in \mathcal{N}_i} \alpha_{i,j}\mathbf{\Theta}\vec{h}_{j}
\end{equation}

Our attention module consists of two $GAT_{SG}$ layers with an MLP and an \gls{elu} activation function in between. The second attention layer generates the final node feature embedding representations ${h^{\prime}}$.
\newline

\textbf{Triple Embedding Learning Module.}
We then generate sets of triples in the form of $\langle \vec{h_i^{\prime}}, \vec{p}^{(kg)}_{k}, \vec{h_j^{\prime}} \rangle$ from the node feature vectors $\mathbf{h^{\prime}}$ in the scene graph and the possible edge feature vectors $\mathbf{p^{(kg)}}$ in the extended scene graph. The model leverages learned triple embeddings to predict the probability of them existing in the corner case ground truth graph.

Conventional methods embed triples by either concatenating or by applying a point-wise dot product operation to the head and tail node feature vectors \cite{Zong2022}. However, these approaches assume homogeneity and do not take into account relational attributes. In our work, we propose a methodology for informing the link prediction using the edge attributes. Extending the edge-weight-based link prediction algorithm proposed by \cite{Zong2022}, we splice the edge weight in between the feature representations of the head and tail nodes for each triple. The spliced feature vector is then passed into a single-layer MLP, followed by a sigmoid function $\sigma$. The output representation corresponds to the probability $\hat{y_k}$ of the triple existing in the corner case graph:
\begin{equation}\label{eq:probability}
    \hat{y_k} = p(\vec{h_i^{\prime}}, \vec{p}^{(kg)}_k, \vec{h_j^{\prime}}) = \sigma \: (MLP \: ( \: \vec{h_i^{\prime}} \: \: || \: \vec{p}_k^{(kg)} \: || \: \vec{h_j^{\prime}} \:))
    \vspace{-0.2cm}
\end{equation}
\newline
\textbf{Loss Function.} We train the model using a binary cross entropy loss function \cite{wang2020comprehensive} between the predicted probabilities $\hat{y_k}$ and the associated ground truth binary values $y_k$. The loss function, $\mathcal{L}$, is defined as follows:
\begin{equation}\label{eq:loss}
    \mathcal{L} = \frac{1}{Q} \sum_{k=1}^Q l_i = \frac{1}{Q} \sum_{k=1}^Q -{(y_k\log(\hat{y_k}) + (1 - y_k)\log(1 - \hat{y_k}))} \;
\end{equation}

\subsection{Scenario Generation.} 
Once the model has generated corner case graphs from regular scene graphs, their entities and relationships are extracted to generate corner case scenarios in simulation. The location and state of traffic agents in the regular driving graphs are used to initialise their start positions while the predicted edges in the corner case graph are used to define the target location of all adversaries in the scenario. We synchronise the adversaries w.r.t. the ego vehicle so that they arrive at the respective predicted locations at the same time as the ego vehicle. These locations can be defined as the predicted lane, the orientation relative to the ego vehicle, or whether they are at a ‘safeDistance’ $(10m)$ or ‘unsafeDistance’ $(1.5m)$.

%===============================================================================
\section{Experimental Results}
\label{sec:result}

\subsection{Quantitative Results}
\begin{figure}[h]
    \centering
    \includegraphics[width=\columnwidth]{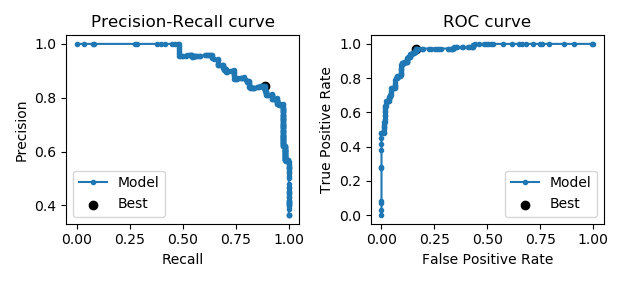}
    \caption{
    Classification evaluation curves over a range of thresholds generated from the model's predictions in the testing dataset. The thresholds correspond to the optimum binary classification thresholds based on the ROC curve.}
    \label{fig:prroc}
    \vspace{-0.2cm}
\end{figure}

\textbf{Prediction accuracy}. We split our dataset into training ($70\%$), validation ($20 \%$), and testing ($10\%$) sets. The model's prediction results on the testing dataset can be seen in Table \ref{table:results} and Figure \ref{fig:prroc}. We compute conventional classification metrics: precision, recall, F1-score, and AUC score.

We then apply Youden's J statistic to find the optimal binary classification threshold and measure the accuracy of our model. 
The best threshold value corresponds to $0.842$ precision and $0.888$ recall which is very close to ideal classification. 
The model achieved a maximum F1-score of $0.865$ using a classification interval of $0.436$. With our method, we achieved a prediction accuracy of $\mathbf{\predacc}$ demonstrating a high level of classification accuracy considering the complexity of the task. 
Our model successfully attained a notable \gls{auc} score of \textbf{0.96} as measured by the ROC curve.
\begin{table}
\vspace{0.2cm}
\centering
\renewcommand{\arraystretch}{1.2}
\begin{adjustbox}{width=0.54\columnwidth}
\begin{tabular}{|l|c|}
\hline
    \textbf{Metric} & \textbf{Score} \\
    \hline
    Prediction Accuracy & \textbf{\predacc} \\
    F1-Score & \fscore \\
    Precision & \precision \\
    Recall & \recall \\
    AUC Score & \textbf{\aucscore} \\
    \hline
\end{tabular}
\end{adjustbox}
\caption{Prediction Results.}
\label{table:results}
\vspace{-0.6cm}
\end{table}

\textbf{Benchmarks}. To validate the fidelity of the generated scenarios, we evaluate the behaviour of $5$ AD systems in simulation: $4$ variations of the CARLA Navigation Agents, and Learning by Cheating \cite{chen2019cheating}, a state-of-the-art method from the CARLA Leaderboard \cite{leaderboard}.
We calculate the Scenario Collision Rate (SCR) \cite{suo2021trafficsim}, as the percentage of collisions, near-misses, and unsafe maneuvers between the ego vehicle and adversaries in the generated scenarios. Given the bounding boxes of the agents present, a collision corresponds to overlapping bounding boxes with $IOU > \epsilon=0.1$. If the ego vehicle is within a threshold of $1.5m$ from the actor, it classifies as a near-miss. For motorway scenarios, an unsafe maneuver is detected when the ego vehicle does not start its journey due to an agent in proximity imposing a hazardous behaviour.  Results are visualised in Table \ref{table:planners}. 

In the majority of scenarios, the baseline methods failed to navigate critical situations and were unable to adjust their path to execute a safe maneuver. All baselines return a higher collision rate than no collision rate in our generated scenarios, demonstrating the effectiveness of our approach in creating challenging situations for autonomous driving systems. Our generated scenarios can be used to extend the CARLA Benchmark to include more unique synthetic corner cases. Overall, our experiments demonstrate the necessity of extensive testing of such systems on critical scenarios like the ones generated from our method, to ensure safe operation.

% This proves the necessity of additional testing in corner

% , proving the necessity of additional testing in corner case scenarios but also demonstrating
% Our results in \ref{table:planners} demonstrate that our model successfully generates critical scenarios for autonomous driving as all our baseline methods return a higher collision than no-collision percentage, proving

% Our method extends the CARLA Benchmark to include more challenging synthetic scenarios.

% To evaluate the fidelity of the generated corner case scenarios, we run $6$ planners and calculate the scenario collision rate (SCR) similar to \cite{feng2022trafficgen}. \ep{That thing that had 100 on carla gets 63perc collisions on ours so we did create more test scenarios for it so it can be improved.}
% \ep{4. Discuss the results}
% Our model successfully generated safety critical scenarios for the planners.
\begin{table}[h]
\renewcommand{\arraystretch}{1.25}
\resizebox{\columnwidth}{!}{
\begin{tabular}{|ll|l|l|l|l|}
\hline
\multicolumn{2}{|c|}{AD Methods}                                                                                           & Collision      & \begin{tabular}[c]{@{}l@{}}No\\ Collision\end{tabular} & \begin{tabular}[c]{@{}l@{}}Near \\ Miss\end{tabular} & \begin{tabular}[c]{@{}l@{}}Unsafe \\ Maneuver\end{tabular} \\ \hline
\multicolumn{1}{|l|}{\multirow{4}{*}{\begin{tabular}[c]{@{}l@{}}CARLA \\ Behaviour \\ Profiles\end{tabular}}} & Basic      & \textbf{58.62} & {\ul 20.69}                                            & 10.34                                                & 10.34                                                      \\ \cline{2-6} 
\multicolumn{1}{|l|}{}                                                                                        & Normal     & \textbf{41.38} & 20.69                                                  & 6.9                                                  & {\ul 31.03}                                                \\ \cline{2-6} 
\multicolumn{1}{|l|}{}                                                                                        & Cautious   & \textbf{41.38} & 20.69                                                  & 3.45                                                 & {\ul 34.48}                                                \\ \cline{2-6} 
\multicolumn{1}{|l|}{}                                                                                        & Aggressive & \textbf{37.93} & 24.14                                                  & 6.9                                                  & {\ul 31.03}                                                \\ \hline
\multicolumn{2}{|l|}{Learning By Cheating*}                                                                                & \textbf{63.12} & {\ul 21.05}                                            & 15.79                                                & 0                                                          \\ \hline
\end{tabular}
}
 \caption{Scenario Collision Rates (SCR) [\%] as performance indicators of baseline AD methods, demonstrating their unreliable performance in our generated corner case scenarios.}
    \label{table:planners}
\end{table}

\subsection{Qualitative Results.} 
\begin{figure}[h]
    \centering
        \includegraphics[width=0.9\linewidth]{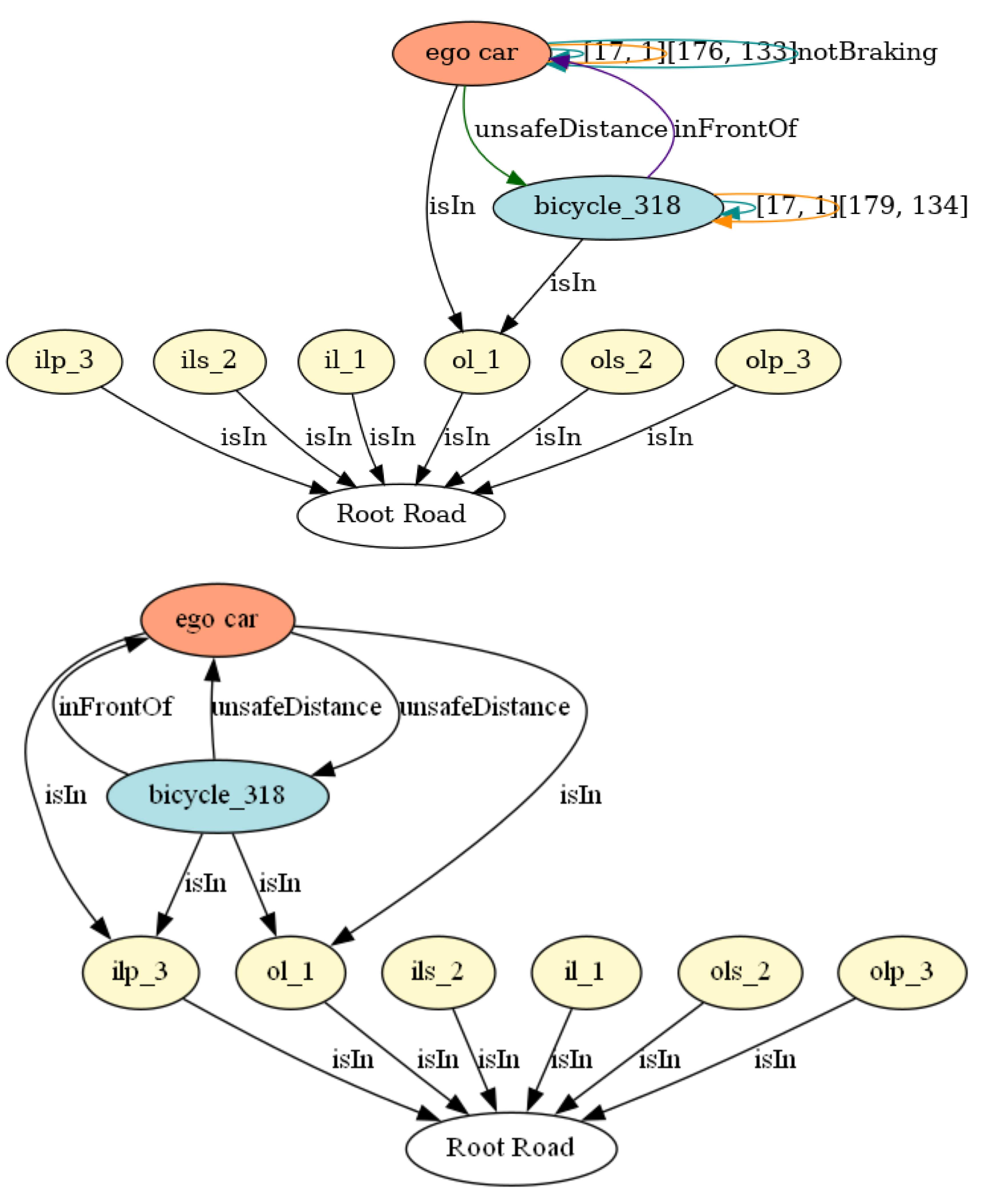}
        \caption{Ground truth (up) and predicted corner case scene graph (down) generated from the regular scenario in Fig. \ref{fig:fig2}. The cyclist is now in the same lane as the ego vehicle and at an unsafe distance. The model has mostly successfully predicted the corner case edges but has generated two corner case scenes in the same graph.}
        %i.e. the cyclist is too close for the ego vehicle to slow down in time, thus imposing a critical scenario.
        \label{fig:qualitative}
        \vspace{-0.3cm}
\end{figure}

Qualitative results from a target and its predicted corner case graph from a scenario in our dataset are visualised in Figure \ref{fig:qualitative}. The model has successfully changed the `toLeftOf' relation to `inFrontOf' and generated an `unsafeDistance' relation, reflecting the ground truth. The model has correctly perturbed the `isIn' relation of the bike to the same lane as the ego vehicle but has predicted this for both the lane and the pavement, hence the model has generated two plausible corner case scenes in the same graph. 
\vspace{-0.3cm}

\section{Ablation Study}
% Our objective is to determine the architecture and channel sizes for our model that yield optimal performance while maintaining a lightweight architecture.

\begin{figure}
    \centering
    \includegraphics[width=0.9\columnwidth]{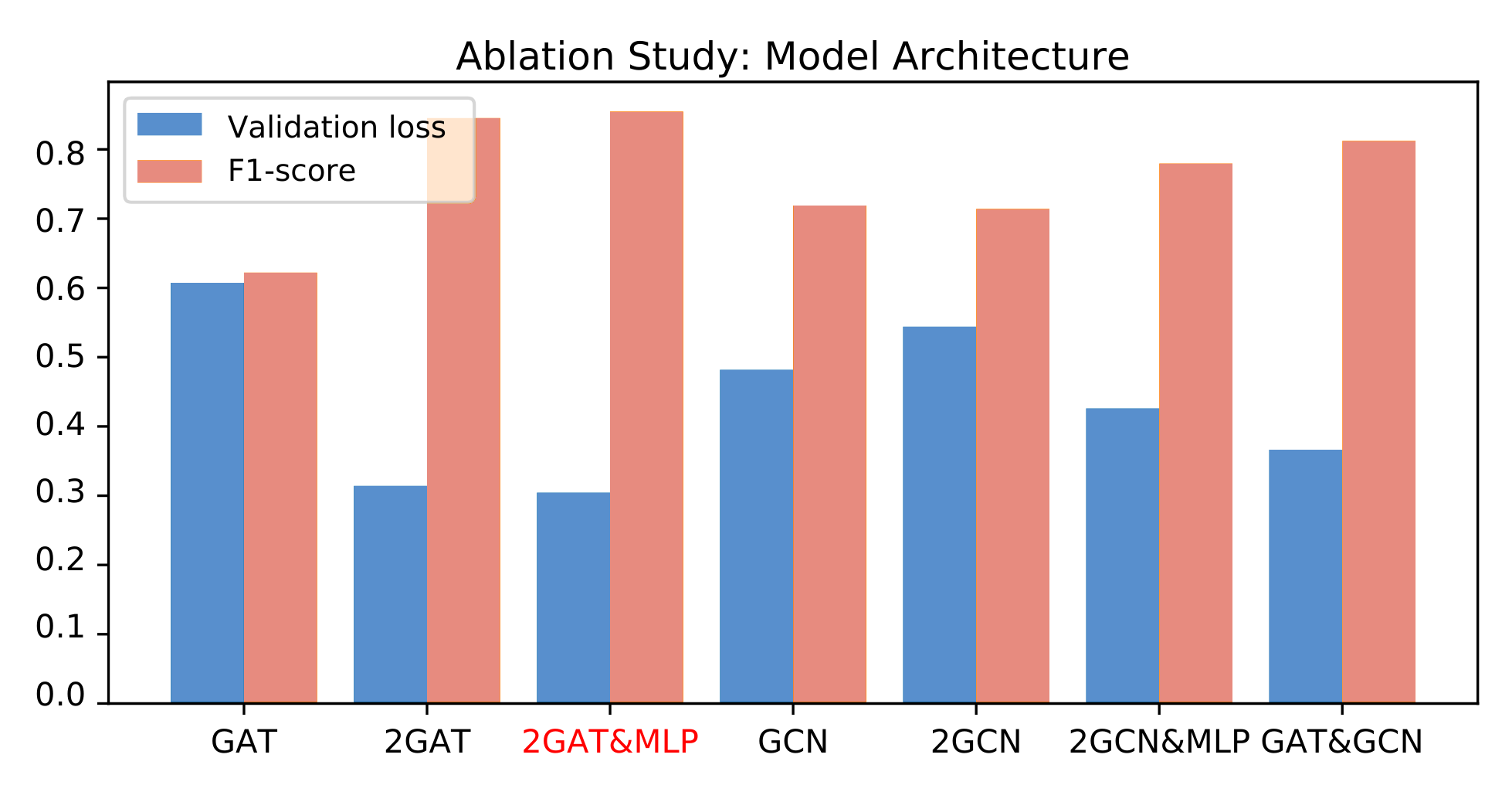}
    \caption{Ablation study to find the optimal architecture that minimises the validation loss and increases the F1-score in our testing dataset.}
    \label{fig:ablation}
            \vspace{-0.5cm}
\end{figure}

\textbf{Architecture.} 
We evaluate different variations between the layers and modules of our model to establish the optimal architecture that maximises the F1-score on the testing dataset while minimising the validation loss. Results are visualised in Figure \ref{fig:ablation}.
The models were evaluated on 3-folds of the dataset using k-fold cross validation splitting the data into k subsets. The models were trained K times, each time using a different subset as the validation set and the remaining K-1 subsets as the training set. The results were then averaged to reduce the bias in the model's performance estimates.

\textbf{Channels.} 
% We tune four parameters: the learning rate ($\mathbf{learning\_rate}$), the hidden channels of the \gls{mlps} ($\mathbf{mlp\_hid}$), the hidden channels of the MLP that predicts the triples ($\mathbf{triple\_mlp\_hid}$), and the channels of the $GAT_{SG}$ ($\mathbf{gat_{sg}\_channels}$). The remaining parameters are either set to 1 to generate flat vector representations, or directly derive from one of the others. 
To identify the key parameters to prioritize during the tuning process, we calculate the \emph{correlation} and \emph{importance} scores for each layer. Correlation derives from the linear correlation between the tunable parameter and the chosen evaluation metric. Importance is calculated after training a random forest using the parameters as inputs and extracting the random forest feature importance values. The results in Table \ref{table:metrics} demonstrate that the most important parameters to tune are the hidden channels of the \glspl{mlp}, as $\mathbf{mlp\_hid}$ scored the largest importance and correlation values. On the contrary, the $\mathbf{learning\_rate}$ returned a relatively strong negative correlation score on the validation loss, even though it had the second highest importance score. The resulting number of channels for the encoding \gls{mlps} can be seen in Table \ref{table:mlp_enc}. The attention module consists of a $GAT_{SG}$[1, 64] followed by a MLP[64, 128, 256], and a $GAT_{SG}$[256, 1]. The hidden channel of the MLP for the triples had a relatively large negative correlation as seen in Table \ref{table:metrics} and thus we used a small MLP[1, 4, 1] to encode the triples' feature vectors.
\begin{table}[h]
\centering
\renewcommand{\arraystretch}{1.3}
\begin{tabular}{c|c|c}
    Parameter & Importance & Correlation \\
     \hline 
     $\mathbf{mlp\_hid}$ & 0.433 & 0.168 \\ 
     $\mathbf{learning\_rate}$ & 0.307 & -0.310\\ 
     $\mathbf{triple\_mlp\_hid}$ & 0.159 & -0.254 \\ 
     $\mathbf{gat_{sg}\_channels}$ & 0.101 & -0.164\\
    \end{tabular}
      \caption{Importance and correlation scores of the parameters w.r.t validation loss.}
    \label{table:metrics}
    \vspace{-0.6cm}
\end{table}

\begin{table}[h]
\centering
\renewcommand{\arraystretch}{1.3}
\begin{tabular}{c|c|c|c|c}
Input & In & Hid & Out & Output\\
     \hline 
     $\mathbf{x}$ & 11 & 64 & 1 & $\mathbf{h}$\\ 
     $\mathbf{edge\_attr}$ & 9 & 64 & 1 & $\mathbf{p}$ \\ 
     $\mathbf{kg\_edge\_attr}$ & 9 & 64 & 1 & $\mathbf{p}^{(kg)}$\\ 
    \end{tabular}
    \caption{MLP channel sizes used to encode node and edge attributes.}
            \vspace{-0.5cm}

    \label{table:mlp_enc}
\end{table}
%===============================================================================

\section{Conclusion}
\label{sec:conclusion}
Our proposed methodology successfully generates corner case scenarios perturbing regular driving scene graphs achieving exceptionally high prediction accuracy across our dataset. We extend conventional \gls{hgnn} attention and triple embedding learning methods to capture the complexity of traffic scenes and predict links corresponding to corner case graphs. Quantitative and qualitative results demonstrate the effectiveness of our approach in generating challenging critical scenarios for baseline AD methods. Overall, our proposed method establishes a strong foundation for leveraging scene graphs and \glspl{gnn} to generate corner case scenarios.

\bibliographystyle{IEEEtran}
\bibliography{biblio}

\end{document}